\documentclass[10pt, a4paper]{article}
\usepackage{lrec}
\usepackage{multibib}
\newcites{languageresource}{Language Resources}
\usepackage{graphicx}
\usepackage{tabularx}
\usepackage{linguex}
\usepackage{soul}

\usepackage{epstopdf}
\usepackage[utf8]{inputenc}

\usepackage{hyperref}
\usepackage{xstring}

\usepackage{color}

\title{Morphological Disambiguation of South Sámi with FSTs and Neural Networks}

\name{Mika Hämäläinen, Linda Wiechetek}

\address{University of Helsinki, UiT The Arctic University of Norway \\
         Finland, Norway \\
         mika.hamalainen@helsinki.fi, linda.wiechetek@uit.no\\}

\abstract{
We present a method for conducting morphological disambiguation for South Sámi, which is an endangered  language. Our method uses an FST-based morphological analyzer to produce an ambiguous set of morphological readings for each word in a sentence. These readings are disambiguated with a Bi-RNN model trained on the related North Sámi UD Treebank and some synthetically generated South Sámi data. The disambiguation is done on the level of morphological tags ignoring word forms and lemmas; this makes it possible to use North Sámi training data for South Sámi without the need for a bilingual dictionary or aligned word embeddings. Our approach requires only minimal resources for South Sámi, which makes it usable and applicable in the contexts of any other endangered language as well. \\ \newline  }

\begin{document}

\maketitleabstract

\section{Introduction}

Sámi languages are a part of the Uralic language family, and like many other Uralic languages they are endangered. The languages of this family are synthetic, meaning that they exhibit a great deal of inflectional and derivational morphology making their processing with computational means far from trivial. 

In this paper, we present a method for morphological disambiguation of South Sámi (ISO 639-3 code sma) by using a morphological FST (finite-state transducer) analyzer and a Bi-RNN (bi-directional recurrent neural network) trained on North Sámi (ISO 639-3 code sme) data and synthetically generated South Sámi data. The disambiguation process takes in all the morphological readings produced by the FST and uses the neural network to pick the contextually correct disambiguated reading. 

North and South are not direct neighbors in the dialect continuum, but share a big part of the lexicon and many grammatical features like an elaborate case system, non-finite clause constructions, a large amount of verbal and nominal derivations.
However, they have a number of distinctions in lexicon, morphology and syntax.

One of the important differences is the omission of the copula verb in South Sámi, but not or less so in North Sámi. The typical word order is SOV (subject-object-verb) in South Sámi, and SVO (subject-verb-object) in North Sámi. 
The case system is slightly different as well. South Sámi distinguishes between inessive (place) and elative (source) case \cite{Bergsland1994}. In North Sámi, this is synthesized in one morpho-syntactic case, called locative case.

In addition to the aforementioned differences, also the homonymies are not the same. In North Sámi, regular noun homonymies are genitive/accusative and comitative singular/locative plural. In South Sámi, on the other hand, they are illative plural/accusative plural and essive (underspecified as regards number)/inessive plural/comitative singular.

\begin{table*}[!htb]
\centering
\begin{tabular}{|l|c|}
\hline
\multicolumn{1}{|c|}{Sentence} & \textbf{\textit{Gos dáppe lea máddi?} `Where is the South here?'} \\ \hline
\hline
\multicolumn{1}{|c|}{FST output} & \begin{tabular}[c]{@{}c@{}}{[}'gos+Adv+Subqst', 'gos+Adv'{]}, {[}'dáppe+Adv'{]}, {[}'leat+V+IV+Ind+Prs+Sg3'{]}, \\ {[}'máddat+V+TV+Imprt+Du2', 'máddat+V+TV+PrsPrc', 'máddi+N+Sg+Nom'{]}, {[}?+CLB{]}\end{tabular} \\ \hline
Source sequence & Adv Subqst \_ Adv \_ IV Ind Prs Sg3 V \_ Du2 Imprt N Nom PrsPc Sg TV V \_ CLB \\ \hline
Target sequence & \begin{tabular}[c]{@{}c@{}}Adv \_ Adv \_ Mood=Ind Number=Sing Person=3 Tense=Pres VerbForm=Fin V \_ \\ Case=Nom Number=Sing N \_ CLB\end{tabular} \\ \hline
\end{tabular}
\caption{An example of the training data}
\label{tab:training}
\end{table*}

Even in the context of morphologically rich languages, a simple POS (part-of-speech) tagging is often not enough as it only reduces some of the ambiguity, and is not enough for lemmatization, for instance. Then again, without lemmatization and the small amount of data available for these languages, modern NLP methods such as word embeddings cannot be as reliably used as in the case of majority languages.

South Sámi, with its estimated number of 500 speakers, is categorized as severely endangered by UNESCO \cite{moseley_2010} and is spoken in Norway and Sweden. The language is spoken in Norway and Sweden and its bilingual users frequently face bigger challenges regarding literacy in the lesser used language than in the majority language due to reduced access to language arenas \cite{Outakoski2015,Lindgren2016}.

The central tools used for disambiguation of Sámi languages are \emph{finite state transducers} and \emph{Constraint Grammars}.
Constraint Grammar is a rule-based formalism for writing disambiguation and syntactic annotation grammars \cite{Karlsson:1990,Karlsson:1995}. Constraint Grammar relies on a bottom-up analysis of running text. Possible but unlikely analyses are discarded step by step with the help of morpho-syntactic context.
The \textit{vislcg3}
implementation\footnote{\url{http://visl.sdu.dk/constraint_grammar.html} (accessed 2018-10-08), also \cite{Didriksen2016}} is used in particular.

South Sámi has several Constraint Grammars including a morpho-syntactic disambiguator, a shallow syntactic analyzer, and a dependency analyzer \cite{Antonsen2010,Antonsen2011}. 
Antonsen and Trosterud (2011) use a fairly small Constraint Grammar (115 rules) for South Sámi part of speech (POS) and lemma disambiguation, resulting in a precision of 0.87 and a recall of  0.98 for full morpho-syntactic disambiguation. 
While these are very good results with a comparatively small workload, they require the work of a linguist with knowledge of the language or a linguist and a language expert in addition. 
However, we want to show how grammatical tools can be built in the absence of these.


Whereas our paper deals with South Sámi disambiguation, the main purpose of this work is to demonstrate that a disambiguator can be built with relatively few resources based on a morpho-syntactically related language. This is useful, not only in the wider context of Sámi languages, but also for other endangered languages as it provides the language community quickly with much-needed resources while there are children - the future speakers - learning the language. Our approach follows the previously established ideology for using FSTs together with neural networks to solve the problem of disambiguation \cite{ens2019morphosyntactic}.

\section{Related Work}

Parallel texts have been used to deal with morphological tagging in the context of low-resourced languages \cite{P16-1184}. They use aligned parallel sentences to train their their Wsabie-based model to tag the low-resource language based on the morphological tags of a more resourced language sentences in the training data. A limitation of this approach is that the morphological relatedness of the high-resource and low-resource languages has to be high.

Andrews et al. \cite{P17-1095} have proposed a method for POS (part of speech) tagging of low-resource languages. They use a bilingual dictionary between a low-resource and high-resource language. In addition, their system requires monolingual data for building cross-lingual word embeddings. The resulting POS tagger is trained on an LSTM neural network, and their approach performs consistently better than the other approaches on the benchmarks they report.

Lim et al. \cite{lim2018multilingual} present an approach for syntactic parsing of Komi-Zyrian and North Sámi data using multilingual word embeddings. They use pre-trained word-embeddings of two high-resource languages; Finnish and Russian. Then they train monolingual word-embeddings for the low-resource languages from small corpora. They project these individual word embeddings into a single space by using bilingual dictionaries for alignment. The parser was implemented as an LSTM based model, and its performance is higher for POS tagging than for syntactic parsing. The most important finding for our purposes is that including a related high-resource language improves the accuracy of their method.

DsDs \cite{plank2018distant} is a neural network based part-of-speech tagger intended to be used in the context of low-resource languages. Their core idea is to use a bi-LSTM model to project POS tags from one language to another with the help of lexical information and word embeddings. Their experiments in a low-resource setting reveal that including word embeddings can boost the model, but lexical information can also help to a smaller degree.

The scope of a great part of the related work is limited to POS tagging. Nevertheless, the morphologically rich Uralic languages call for a more full blown morphological disambiguation than a mere POS tagging in order to make higher-level NLP tools usable for these languages. Moreover, our approach cannot count on the existence of high-quality bilingual dictionaries between morphologically similar languages nor aligned word embeddings, as such resources are not easily available for endangered languages.

\section{Data and Tools}

The training data for South Sámi disambiguation comes from the Universal Dependencies Treebank of the related North Sámi language \citelanguageresource{sheyanova:2017}. Out of all the Sámi languages, North Sámi has by and large the biggest amount of NLP resources available and therefore its use as a starting point for related languages makes perfect sense. The treebank consists of 26K tokens and comes pre-divided into a training and testing datasets.

In addition to the treebank, we use FSTs for both North Sámi and South Sámi with UralicNLP \citelanguageresource{uralicnlp_2019}. These transducers are integrated in the open \textit{GiellaLT} infrastructure \citelanguageresource{Moshagen2014} for Uralic languages. The FSTs take in a word in an inflectional form and produce all the possible morphological readings for it.

In order to evaluate our system, we use a small dataset for South Sámi that has been disambiguated automatically by a Constraint Grammar and checked manually. Currently, the dataset is not publicly available. The data consists of 1994 disambiguated sentences and we only use it for the evaluation.

\begin{table}[!htb]
\centering
\begin{tabular}{|c|c|c|}
\hline
 & \textbf{North Sámi} & \textbf{South Sámi} \\ \hline
Average & 3.1 & 1.8 \\ \hline
\end{tabular}
\caption{Average ambiguity}
\label{tab:ambiguity}
\end{table}

Table \ref{tab:ambiguity} shows the average morphological ambiguity in the North Sámi training set and South Sámi test set when the FSTs are used to produce all morphological readings for every word in the corpus. As we can see, North Sámi exhibits a much higher degree of morphological ambiguity than South Sámi.

For generating more data, we use the South Sámi lemmas from the South Sámi-Norwegian dictionary located in the \textit{GiellaLT} infrastructure \citelanguageresource{Moshagen2014}. The dictionary has 11,438 POS tagged South Sámi lemmas. We only use this dictionary for South Sámi words and omit all the Norwegian translations in our method.

\begin{table*}[htb]
\centering
\begin{tabular}{|c|c|}
\hline
\textbf{Template} & \textbf{Target morphology} \\ \hline
\hline
(N Sg Nom) (N Sg Ill) (V IV Ind Prs Sg3) & \begin{tabular}[c]{@{}c@{}}(N Case=Nom Number=Sing) (N Case=Ill Number=Sing) \\ (V Mood=Ind Number=Sing Person=3 Tense=Pres VerbForm=Fin)\end{tabular} \\ \hline
(N Sg Nom) (Adv) (V TV Ger) & (N Case=Nom Number=Sing) (Adv) (V VerbForm=Ger) \\ \hline
(N Sg Nom) (N Sg Ine) & (N Case=Nom Number=Sing) (N Case=Ine Number=Sing) \\ \hline
mannem (N Sg Acc) (V TV Ind Prs Sg1) & \begin{tabular}[c]{@{}c@{}}(Pron Case=Acc Number=Sing Person=1 PronType=Prs) \\ (N Case=Acc Number=Sing) \\ (V Mood=Ind Number=Sing Person=1 Tense=Pres VerbForm=Fin)\end{tabular} \\ \hline
(N Sg Nom) (N Sg Ela) (V IV Ind Prs Sg3) & \begin{tabular}[c]{@{}c@{}}(N Case=Nom Number=Sing) (N Case=Ela Number=Sing)\\ (V Mood=Ind Number=Sing Person=3 Tense=Pres VerbForm=Fin)\end{tabular} \\ \hline
altemse (V TV Ind Prs Sg1) (N Ess) & \begin{tabular}[c]{@{}c@{}}(Pron Case=Acc Number=Sing Person=3 PronType=Prs) \\ (V Mood=Ind Number=Sing Person=1 Tense=Pres VerbForm=Fin)  \\ (N Case=Ess)\end{tabular} \\ \hline
\end{tabular}
\caption{Templates for generating South Sámi data}
\label{tab:generate}
\end{table*}

\section{Neural Disambiguation}

We train a sequence-to-sequence Bi-RNN model using OpenNMT \cite{opennmt} with the default settings except for the encoder where we use a BRNN (bi-directional recurrent neural network) instead of the default RNN (recurrent neural network) as BRNN has been shown to provide a performance gain in a variety of tasks. We use the default of two layers for both the encoder and the decoder and the default attention model, which is the general global attention presented by Luong et al. \cite{luong2015effective}. 

We experiment with two models, one that is trained with the North Sámi Treebank only, and another one that is trained with South Sámi text generated by templates and the North Sámi data. Both models are trained for 60,000 training steps with the same random seed value.

The North Sámi data gives us the target sequence, that is the correct morphological tags and the POS tag. However, the source sequence has to be generated automatically before the training. For this, we use the North Sámi FST analyzer. We produce all the possible morphologies for each word in the Treebank. The training is done from a sorted list of homonymous readings for each word separated by a character indicating word boundary to the disambiguated set of homonymous readings from the UD (universal dependencies) TreeBank on a sentence level. In other words, the only thing the model sees are morphological tags on the source and the target side. Lemmas and words are dropped out so that the model can be used for South Sámi without the need of aligned word embeddings or dictionaries. This is illustrated in Table \ref{tab:training}.

For producing synthetic data, we wrote six small templates that reflect some common morpho-syntactic differences between South Sámi and North Sámi. This is, for instance, the absence of elative and inessive case in North Sámi, both of which are merged into the single locative case. For each template, we produce 20 different ambiguous sentences by selecting words fitting to the template at random from the South Sámi dictionary and inflecting them accordingly with the FST. Once the words are inflected, we can analyze them to get the ambiguous reading. The templates can be seen in Table \ref{tab:generate}.

\section{Results and Evaluation}

We evaluate the models, the one trained only with the North Sámi Treebank and the one that had additional template generated training data, with the disambiguated gold standard that exists for South Sámi. As the South Sámi gold standard follows the \textit{GiellaLT} FST tags, we converted the tags automatically into UD format, since the neural network is trained to output UD tags.

The evaluation results are shown in Table \ref{tab:results}. The first column shows the percentage of sentences that have been fully disambiguated correctly, the second columns shows this on a word level i.e. how many words were fully correctly disambiguated and finally the last column shows the accuracy in POS tagging. The results indicate that adding the small synthetically generated data to the training boosted the results significantly. 

\begin{table}[h!]
\centering
\begin{tabular}{|l|c|c|c|}
\hline
 & \begin{tabular}[c]{@{}c@{}}\textbf{Fully correct} \\ \textbf{sentences}\end{tabular} & \begin{tabular}[c]{@{}c@{}}\textbf{Fully correct} \\ \textbf{words}\end{tabular} & \textbf{POS} \textbf{correct} \\ \hline
\begin{tabular}[c]{@{}l@{}}N. Sámi\\ only\end{tabular} & 12.0\% & 37.6\% & 59.7\% \\ \hline
\begin{tabular}[c]{@{}l@{}}N. Sámi \&\\  templates\end{tabular} & 13.0\% & 42.2\% & 66.4\% \\ \hline
\end{tabular}
\caption{Evaluation results of the two different models on South Sámi data}
\label{tab:results}
\end{table}

As for the incorrectly disambiguated morphological readings, there is a degree to how incorrect they are. This is shown in Table \ref{tab:results-errors}, which shows the errors based on how many morphological tags were predicted wrong. In both cases, more than half of the wrongly disambiguated words only differ by one tag from the gold standard. The results for the model trained on the additional template data show that the errors the model makes are still closer to the correct reading.

\begin{table}[h!]
\centering
\begin{tabular}{|l|c|c|l|c|}
\hline
 & \multicolumn{1}{l|}{\textbf{1 tag}} & \multicolumn{1}{l|}{\textbf{2 tags}} & \textbf{3 tags}& \multicolumn{1}{l|}{\textbf{more tags}} \\ \hline
\begin{tabular}[c]{@{}l@{}}North Sámi \\ only\end{tabular} & 58.4\% & 11.8\% & 15.8\% & 14.0\% \\ \hline
\begin{tabular}[c]{@{}l@{}}North Sámi \\ \& templates\end{tabular} & 60.8\% & 12.1\% & 15.1\% & 12.0\% \\ \hline
\end{tabular}
\caption{Errors based on the number of erroneous morphological tags on a word level}
\label{tab:results-errors}
\end{table}

Below, we are having a closer look at the actual sentences and their analyses, shedding some light on the shortcomings the neural network and suggesting improvements.
In ex. \ref{daelie}, our system erroneously picks the nominal singular nominative instead of the adverb reading for \textit{daelie}. The nominal reading, however, is very rare.

\exg. Daelie dle geajnam gaavnem!\label{daelie}\\
now;then\textsc{sg.nom} so street\textsc{.acc} find\textsc{.prs.sg1}\\
`Then I find the street!'





Negation verbs pose a problem to the neural network. Of the 238 instances only very few negation verbs - despite not being homonymous with any other forms - are analyzed as such. In ex. \ref{im}, \textit{im} `I don't' is analyzed a an indicative past tense verb 1st person singular (the last of which is correct) despite the fact that \textit{im} is not ambiguous. 

\exg. Im	sïjhth	gåabph	gih,	men	tjidtjie	jeahta	månnoeh.\label{im}\\
not\textsc{.neg.sg1} want\textsc{.conneg} anywhere then, but mother says us\textsc{.du1.nom}\\
`I don't want anywhere then, but mother says us two will go.'

There are other difficulties related to negation in the system.
In the following example, the neural network predicts more tokens than the sentence contains, i.e. a negative verb (correctly) and a connegative form (erroneously) usually preceded by the negative verb.

\exg. -	Aellieh!\label{aellieh}\\
- not\textsc{.neg.imprt.sg2}\\
`- Don't!'



\section{Discussion and Conclusions}

Uralic languages  are highly ambiguous in terms of their morphology, and the linguistic resources such as annotated corpora for these languages are quite limited. This poses challenges in the use of modern NLP methods that have been successfully employed on high-resource languages.  In order to overcome these limitations, we proposed a representation based on  the ambiguous morphological  tags of  each word in a sentence.

We have presented a viable way of disambiguation for South Sámi based on an FST and training data on North Sámi with minimal templates needed to cover some of the morpho-syntactic differences of the two languages. The preliminary results look promising, especially since there are nine different Sámi languages. Not to mention similar situations for other endangered languages, where data for a similar language is available.

Our method is more of a hybrid pipeline of rule-based FSTs that produce the possible morphological readings and a neural network that does the disambiguation. This makes it possible to replace the FST with some other rule-based solution or a neural network based morphological analyzer, given that recent research has shown promising results for the use of neural networks in morphology of endangered languages \cite{schwartz-etal-2019-bootstrapping,silfverberg-tyers-2019-data}.

Moreover, our pipeline can be further enhanced by rules. In our experiments, we had the neural network disambiguate out of all the possible morphological readings. Instead of doing that, it is possible to disambiguate first with a rule-based tool such as a Constraint Grammar, and use the neural network to disambiguate the remaining ambiguity. That way we do not need to guess what we already know. It is particularly important to make sure that if the morphology is known, the neural network would not be used to guess it again. This would allow for combining the best of the two worlds; the accuracy of the rule-based methods and the scalability of a neural network.

An interesting question for the future is how far one could get in disambiguation with our proposed method if one was only to train the model by using templates. As even a small number of templates was enough to improve the results noticeably, an entirely template based approach does not seem to be entirely out of the question. Especially if the templates were constructed with more generative freedom such as by following a formalism deriving from CFG (context-free grammar). The use of synthetically generated source data is known to improve NMT (neural machine translation) models when the target data is of a high quality (see \cite{sennrich-etal-2016-improving}). Also, some promising work has been conducted in fully synthetically generated parallel data in NMT \cite{hamalainen2019template}.

This year has been particularly good for Uralic languages with small UD Treebanks recently published for Skolt Sámi, Karelian, Livvi, Komi-Permyak and Moksha. This means that in the future we can try different variations of our method with these languages as well with minimal modifications to the current approach as all of these languages have rule-based FSTs available in the \textit{GiellaLT} infrastructure.

\section{Acknowledgments}
We would like to thank Lene Antonsen and Anja Regina Fjellheim Labj for their work on the South Sámi Constraint Grammar disambiguator within the \textit{GiellaLT} infrastructure and for making their automatically annotated and manually corrected South Sámi corpus available to us.

\section{Bibliographical References}\label{reference}

\bibliographystyle{lrec}
\bibliography{lrec2020W-xample-kc}

\section{Language Resource References}
\label{lr:ref}
\bibliographystylelanguageresource{lrec}
\bibliographylanguageresource{languageresource}

\end{document}